%% file: acl_latex.tex
\pdfoutput=1

\documentclass[11pt]{article}

\usepackage[final]{acl}

\usepackage{times}
\usepackage{latexsym}
\usepackage{booktabs}
\usepackage{amsmath}
\usepackage{multirow}
\usepackage{url}

\usepackage[T1]{fontenc}

\usepackage[utf8]{inputenc}

\usepackage{microtype}
\usepackage{xspace}

\usepackage{inconsolata}

\usepackage{graphicx}
\newcommand{\ours}{{\scshape RuleR}\xspace}

\title{\ours: Improving LLM Controllability by Rule-based Data Recycling}

\author{ 
  \textbf{Ming Li}\textsuperscript{*1}, 
  \textbf{Han Chen}\thanks{Equal Contribution.}, 
  \textbf{Chenguang Wang}\textsuperscript{*2}, 
  \textbf{Dang Nguyen}\textsuperscript{1}, 
  \textbf{Dianqi Li}, 
    \textbf{Tianyi Zhou}\textsuperscript{1}\\
  \textsuperscript{1}University of Maryland \textsuperscript{2}Stony Brook University\\
  \{minglii, tianyi\}@umd.edu \\
  Project: \url{https://github.com/tianyi-lab/RuleR}
}

\begin{document}
\maketitle

\begin{abstract}
Despite the remarkable advancement of Large language models (LLMs), they still lack delicate controllability under sophisticated constraints, which is critical to enhancing their response quality and the user experience. 
While supervised fine-tuning (SFT) can potentially improve LLM controllability, curating new SFT data to fulfill the constraints usually relies on human experts or proprietary LLMs, which is time-consuming and expensive. 
To bridge this gap, we propose \textbf{\underline{Rule}-based Data \underline{R}ecycling (\ours)}, a human/LLM-free data augmentation method incorporating multiple constraints into the original SFT data. 
Instead of creating new responses from scratch, \ours integrates linguistic or formatting rules into the original instructions and modifies the responses to fulfill the rule-defined constraints. 
Training on the ``recycled'' data  
consolidates LLM capability to generate constrained outputs, improving LLM controllability while maintaining promising general instruction-following capabilities.  
\looseness-1
\end{abstract}

\section{Introduction}
\input{figure_method}

Despite the remarkable advancement of the current Large language models (LLMs) and the continuous efforts to build high-quality supervised fine-tuning (SFT) datasets, one critical challenge is to generate responses better interacting with humans, with the utility and effectiveness maximized for end-users~\cite{Liu_2024, huang2024trustllm, huang2023trustgpt}. According to the systematic investigation from \citet {Liu_2024}, it is essential for LLMs to constrain their outputs to follow user-specified formats or characteristics. 
In various practical applications, free-formed responses are not legal or directly applicable without any constraint or format being enforced. 
It has also been verified on LLM Agents \cite{li2024formal, chen2023fireact, chen2024agentflan, zhang2024agentohana} that enforcing predefined formats is necessary for tasks. 

However, existing SFT datasets are mainly composed of general instructions without user-specified constraints~\cite{wei2022finetuned, wang-etal-2022-super, alpaca, xu2023wizardlm, zhou2023lima, Li2023ReflectionTuningDR, zhang2023instruction, Xu2024ASO} and thus result in models lacking delicate controllability of the lengths and format of responses~\cite{chen2024benchmarking, xia2024fofo}.  
To enhance the utility of existing SFT data in improving the controllability of LLMs, a potential method is to rewrite or modify instructions and responses by experts such as humans/LLMs~\cite{xu2023wizardlm, Li2023ReflectionTuningDR, Li2024SelectiveRS, Li2024CanLS, he2024complex, dong2024self, wu2024unigen} in order to make them fulfilling multiple constraints, as shown in Figure \ref{fig:method} (top). 
However, the curation of new data is not only costly and inefficient, requiring careful editing by human experts or proprietary LLMs, but also represents a waste of previous efforts: It is impractical to discard all existing data and create brand new data every time we need to add more constraints to the instructions. 
Hence, we raise the question: \textit{\textbf{Can we ``recycle'' existing SFT data without human/LLM editing and enforce various types of constraints in order to improve LLM controllability?}} 
\looseness-1

Drawing inspiration from IFEval \cite{zhou2023instructionfollowing}, which utilizes verifiable constraints to evaluate LLMs' controllability, and the human/model-free data augmentation in Mosaic-IT \cite{Li2024MosaicIE},  
we propose \textbf{\underline{Rule}-based Data \underline{R}ecycling (\ours)}, which automatically ``recycles'' existing SFT data for improving LLM controllability.
As illustrated in Figure \ref{fig:method} (bottom), the key insight of \ours is to automatically build constraint-augmented SFT datasets \textbf{at no cost of human/LLM efforts}, by applying predefined rules to the original instructions and responses. 
Specifically, we manually inspect and construct a diverse set of rules as constraints, which specify the linguistic or formatting constraints on different parts of the response. 

Our predefined rules cover a wide range of diverse constraints generalizable to many application scenarios, ranging from high-level constraints, e.g., controlling the word frequency in the response, to lower-level constraints, e.g., setting specific wrapping formats of some keywords.  
Each rule is composed of (1) multiple templates to produce additional instructions enforcing the constraints,  
and (2) a piece of code that alternately edits the original instruction and response in order to make the edited response fulfill all the constraints appended to the instruction. 
For each sample from the original dataset, we randomly draw several rules to be applied to the editing. This produces an augmented sample with the constraints enforced so it can be used for controllability tuning. The complete list of rules and descriptions can be found in Appendix~\ref{sec:rules}.

Illustrative examples are provided in Figure \ref{fig:toy_example}, which showcase (a) a rule that constrains the number of letters; (b) a rule that specifies the case of specific words (if presenting in the response); and (c) a rule that specifies the wrapping format of specific words. 
To ensure the consistency between the input constraints and the output response, we modify both the instruction and response based on the characteristics of the original data sample. For each sample, we only sample from the rules applicable to the original response, hence avoiding the potential incorrectness of the edited responses. 

\input{figure_example_mini}

Extensive experimental results on the IFEval benchmark on various base models and datasets demonstrate the effectiveness of \ours in enhancing LLM controllability without extra help from humans/models. 
On the other hand, \ours still preserves the general instruction-following ability promoted by the original SFT dataset. This is demonstrated by the instruction-following metrics (Pair-wise comparison and Open LLM Leaderboard). To the best of our knowledge, \ours is the \textbf{first human/model-free data augmentation and recycling approach designed to improve LLM controllability} under multiple constraints. 

\section{Methodology}

\subsection{Preliminaries}   

Given a supervised finetuning dataset $D$, there are $N$ data samples, each represented by a tuple $(x_i, y_i)$, where $x_i$ represents the instruction and $y_i$ represents the corresponding response. Let $p_\theta(\cdot)$ denote the LLM with parameters $\theta$ to be trained. In the instruction tuning setting, $p_\theta$ is typically fine-tuned by maximizing the following objective on all the $N$ samples as $(x_i,y_i)$, in which $y_{i,j}$ represents the $j_{th}$ token of response $y_i$, $ y_{i,<j}$ represents the tokens before $y_{i,j}$, and $l_i$ represents the token length of $y_i$: 
\looseness-1
\vspace{-5mm}

\begin{align}
   \max_\theta \sum_{i=1}^N\sum_{j=1}^{l_i} \log p_\theta\left(y_{i,j} | x_i, y_{i,<j} \right),
\end{align}

\subsection{Rule-based Data Recycling (\ours)}

\subsubsection{Rule Construction}

While most existing methods still require human experts or strong teacher LLMs to generate new data (Figure~\ref{fig:method} (top)), we aim at ``recycling'' instructions and responses from existing SFT datasets to build controllability-focused datasets, without expensive and time-consuming supervision from humans or LLMs. In the following, we introduce a rule-based approach ``\textbf{\underline{Rule}-based Data \underline{R}ecycling (\ours)}'' to create high-quality augmented data for improving LLM controllability. 

\ours reformulates the original instructions and responses by applying rule-based edits according to pre-defined constraints. 
However, not every constraint is applicable to a randomly selected sample without fully rewriting. Hence, we only incorporate constraints compatible with the original sample and those can be implemented with simple rectifications like regular expressions. 
Specifically, we focus on the characteristics of responses that can be defined by rules, e.g., by checking the 220 distinct linguistic features in the LFTK package~\cite{lee-lee-2023-lftk}. 
In addition, we collect constraints from existing works and widely used instruction-tuning datasets. 
These diverse characteristics include punctuation-, word-, sentence-, paragraph-level occurrences, and frequencies.  
Thus we construct rules constraining or specifying the characteristics of the original response. 
In conjunction with the rules containing these characteristics, we also create rules specifying the format of responses to improve LLM's format-following capability. 
\looseness-1

To ensure that the rules selected or sampled for each sample are applicable, we apply the following additional protocols: 
(1) The rules need to be applicable to the original sample. For instance, the rule ``\textit{Generating a title before giving the response}'' is not applicable as we can not generate a title without the help of humans or other additional models. 
(2) The rules should not include removing the content of the original response. For instance, the rule ``\textit{Ensure the word {xxx} is not shown in the response}'' is not applicable since we can not directly remove this word from the response as the removal might disrupt the original semantic integrity. 
(3) The rules should be compatible with the original sample. If the rule is ``\textit{Ensure there are more than {N} sentences in the response}'', then it can not be applied to samples whose responses have $<N$ sentences. 
The complete list of rules is provided in Table~\ref{tab:list}, which covers both high-level constraints such as the term frequency, and lower-level constraints such as specific wrapping formats. 

\subsubsection{Rule Implementation}
\label{sec:rule}

To implement each rule to original instructions and responses, we notate each pre-defined rule as a tuple for simplicity, 
$(\mathbf{S}_k, f_k, g_k)$, where $\mathbf{S}_k$ represents the set of manually curated instruction templates for creating instructions of the $k^{th}$ rule, while $f_k$ and $g_k$ are the corresponding functions to reformulate instructions and responses (if necessary), respectively. 

Specifically, the function $f_k$ selects an appropriate template of a rule for a given data sample and augments the original instruction with an instruction generated by the template. 
It first selects a subset of rules applicable to the characteristics of the response (e.g., presence of keywords, number of sentences, etc.). 
Then, it randomly draw one rule out of the subset and create a formatted instruction of the rule from the template. 
Such rule instruction is appended to the original instruction. 

Specifically, for each data sample $(x_i, y_i)$, the augmented instruction $x_{i,aug}$ is reformulated according to the characteristics of the original sample and the corresponding template sets, i.e.,
\begin{equation}
    x_{i,aug} = f_k(x_i, y_i, \mathbf{S}_k),
\end{equation}

The function $g_k$ is designed to modify the response to be consistent with the augmented instruction (after applying function $f_k$), i.e., with the rule applied. 
Applying $g_k$ either preserves the original response or revises it.   
Some rules do not require editing of responses, e.g., keyword appearance, the number of nouns, etc. For rules defining the case or format of certain parts in the response, modifications are needed. 
The detailed descriptions of each predefined rule can be found in Appendix~\ref{sec:rules}. 
Specifically, the augmented response $y_{i,aug}$ will be optionally modified based on the selected rule $k$:
\begin{equation}
    y_{i,aug} = g_k(x_i, y_i, \mathbf{S}_k). 
\end{equation}
With the augmented sample $(x_{i,aug}, y_{i,aug})$, the training objective becomes:
\vspace{-2mm}
\begin{align}
   \max_\theta \sum_{i=1}^N\sum_{j=1}^{l_{i,aug}} \log p_\theta\left(y_{i,aug,j} | x_{i,aug}, y_{i,aug,<j} \right),
\end{align}
where $y_{i,aug,j}$ represents the $j$th token of response $y_{i,aug}$ and $l_{i,aug}$ represents its token length.

\subsubsection{Multi-rule Implementation}

To create more complex, diverse, and challenging samples, we can extend the previous process to multiple rules randomly drawn from the feasible set of rules. 
We provide examples of multi-rule augmentation in Appendix \ref{sec:examples}, which forces LLMs to learn to follow multiple constraints. 

The detailed experimental setup can be found in Appendix \ref{sec:experimental}, including Implementation Details, Training Datasets, and Evaluation Metrics. 
\looseness-1

\input{table_main_1}

\section{Experimental Results}

\subsection{Main Results}

The main experimental results are presented in Table \ref{tbl:main_1}, containing the performance comparison on the Instruction Following Eval \cite{zhou2023instructionfollowing}, Pair-wise Comparison Winning Score, and the Open LLM Leaderboard \cite{eval-harness}, on 3 different base models and several different instruction tuning datasets. The Pair-wise Comparison Winning Score is calculated as \textit{(Num(Win)$-$Num(Lose))$/$Num(All) $+ 1$} and the values that are greater than $1.0$ represent better responses generated.
Detailed descriptions of evaluation metrics can be found in the Appendix \ref{sec:eval}. 

Compared to the baseline, our method has consistent improvements on the IF Eval benchmark, across different base models and datasets, which aims at measuring LLMs' constraint-following abilities by using verifiable instructions. It is astonishing that our method can improve the IF Eval scores by approximately $10\%$ on some of the configurations, by just utilizing the rule-based recycling method on the original data, without any human/model edition. 
Moreover, the performances keep being positive on originally diverse and high-quality datasets like Recycled WizardLM \cite{Li2023ReflectionTuningDR} and DEITA \cite{liu2023makes}, which further verify the potential of our method. 
Compared with existing methods which enhance LLMs' constraint controllability by generating totally new data, our method focuses more on fully utilizing the potential of existing data. 
\looseness-1

Furthermore, our method not only improves the constraint controllability but also keeps the general instruction-following ability of the original data. The Pair-wise comparison and Open LLM leaderboard results showcase comparable or sometimes better performances compared with the baseline models. 
We hypothesize that the additional constraints largely complicate the original instructions, as shown in Figure \ref{complex_example}, thus forcing the LLMs to understand each constraint before generating responses, thus leading to potentially improved instruction-following abilities.
\looseness-1

\subsection{Ablation Studies}

\input{table_ablation}

In this section, ablation experiments are conducted on Mistral-7B with the Alpaca-GPT4 dataset, aiming to evaluate the impact of several factors.

\textbf{Effect of Templates:} 
As shown in Table \ref{tbl:ablation}, ``Single Temp.'' represents utilizing only one rule-instruction template for each rule, while ``Diverse Temp.'' represents utilizing several different templates, approximately $10$, with the same meaning for each rule and those templates are randomly sampled during the augmentation. 
``Diverse Temp.'' demonstrates superior performance in both IF Eval and Pair-wise winning scores, with slightly lower accuracy on the Open LLM leaderboard. 
This result suggests that ``Diverse Temp.'' enhances the model's constraint controllability while enhancing its general instruction-following capabilities compared to ``Single Temp.'' 
On the contrary, when fixing the rule templates to one single template, potential overfitting to the template might occur and thus negatively influence the model performances.

\textbf{Effect of Rule Numbers:} 
``Max Rule = $x$'' represents the setting in which at most $x$ different rules can be sampled and utilized on each original data sample. 
In the augmentation process, a random value will be sampled in the range of $[0, x]$ as the number of rules in this sample. 
The examples in Figure \ref{fig:toy_example} showcase the scenario when only one rule is implemented and examples in Figure \ref{complex_example} showcase the scenario when multiple rules are implemented. 
Compared to the baseline, nearly all settings show performance improvements across the three evaluation metrics. However, the IF Eval score initially increases, reaching its peak when ``Max Rule = $3$'', before declining. 
The Pair-wise score, on the other hand, consistently decreases from ``Max Rule = $1$'' to ``Max Rule = $5$''. 
These results suggest that applying too many rules to a single sample may impair the LLM's capability, even though sampling and applying multiple rules can be beneficial when done in moderation, which might be because the original instruction becomes so complex.
\looseness-1

\textbf{Effect of Augmentation Rate:} 
``Aug Rate = $x$'' represents there is a probability of $x$ to apply our augmentation to each sample. 
It is observed that as the augmentation rate increases, performance improves on the IF Eval and mostly improves on the Pair-wise evaluation. 
This phenomenon indicates that increasing the augmentation rate primarily enhances the LLM's constraint controllability, and it also has a positive impact on its general instruction-following capability. 
However, the gaps on IF Eval are much larger than the other 2 metrics, indicating this rate will mostly influence the multi-constraint controllability of LLM.

\noindent
\textbf{Detailed Sub-Category Analysis:} 
The detailed sub-category analysis can be found in Appendix \ref{sec:sub}.

\vspace{-2mm}
\section{Conclusion}
\vspace{-2mm}

In this work, we proposed \textbf{\underline{Rule}-based Data \underline{R}ecycling (\ours)}, which modified the original instructions and responses from an existing dataset by 
rule-defined constraints. 
The ``recycled'' data aims to enhance the LLMs' capability to generate outputs fulfilling the constraints specified in the input, 
thereby improving the controllability of LLMs. 
\ours took the first step of exploring rule-based data recycling, 
which can serve as a plug-and-play and easy-to-use method that converts any existing SFT datasets to new datasets for better controllability. 
\looseness-1

\section*{Limitations}
Our method focuses on improving LLM controllability by rule-based editing of existing data, thereby avoiding the extra cost of data generation by humans or expert models.
Though it saves the cost of human/model editing, the rules inevitably limit the types of constraints that can be applied to modify the original data. 
In the presented \ours, all the constraints and rules are based on verifiable shallow syntactic characteristics such as the occurrence and frequency of words or sentences while lacking constraints and controllability on the semantic feature or content. 
This implies a potential of \ours to be further enhanced by modifying the semantic content with a smaller model, which retains a comparable efficiency of rule-based editing. 
\bibliography{custom}

\clearpage

\appendix

\clearpage
\input{appendix_experimental}

\clearpage
\input{appendix_examples}

\clearpage
\input{appendix_sub}


\clearpage
\input{appendix_evaluation}

\clearpage
\input{appendix_rules}

\end{document}

%% file: figure_method.tex
\begin{figure}[!t]
    \centering
    \includegraphics[width=1\columnwidth]{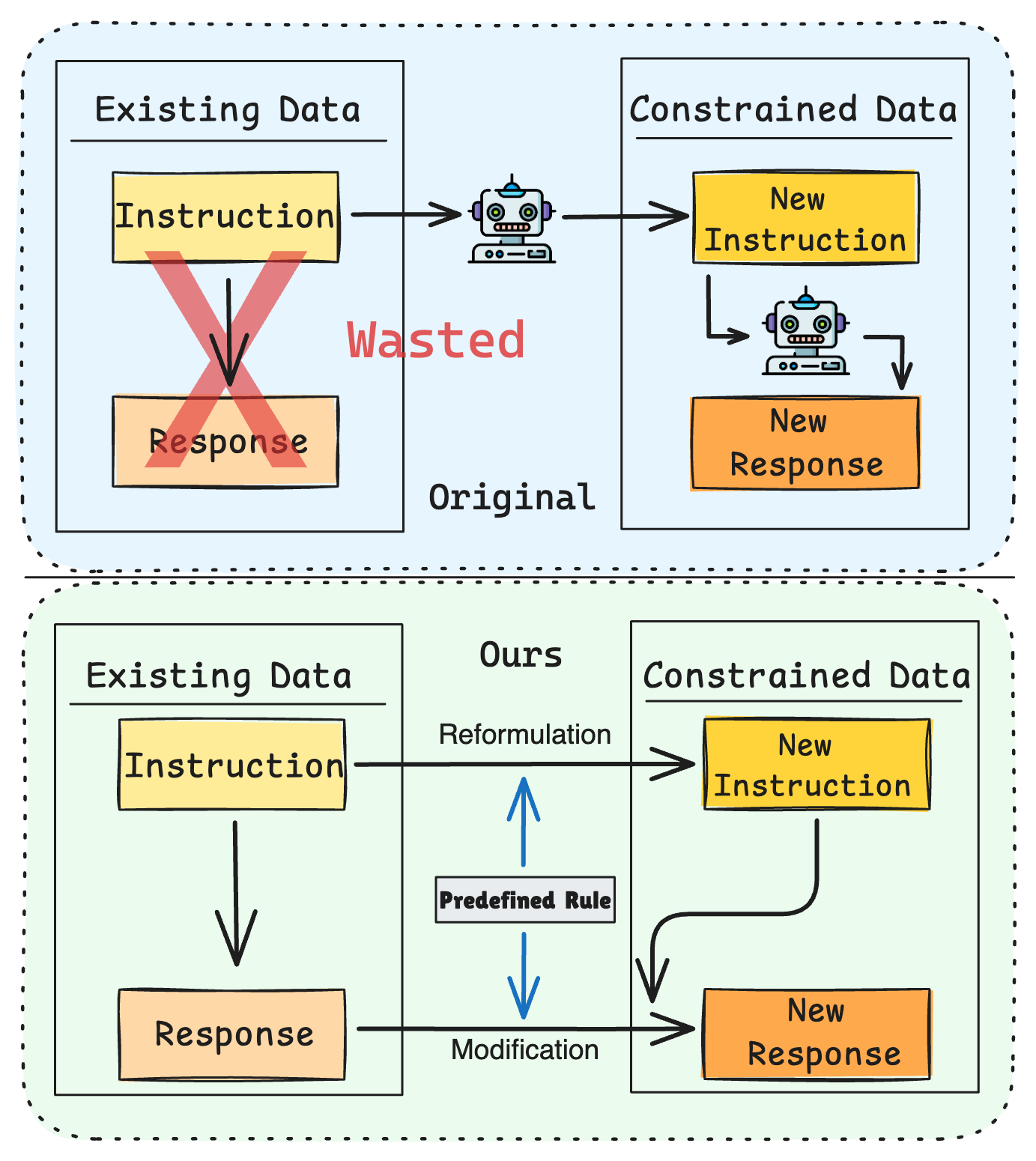}
    \caption{Comparing widely-used data generation strategy (top) and \ours (bottom) enhancing LLM controllability. Most existing methods rely on human/model rewriting to generate new instructions and responses. However, discarding existing data is a waste of effort. Our \ours demonstrates that simple rule-based (human/model-free) editing of existing data can generate new SFT data that improves LLM controllability.}
    \label{fig:method}
    \vspace{-6mm} 
\end{figure}

%% file: figure_example_mini.tex
\begin{figure}[t]
    \centering
    \includegraphics[width=1.0\linewidth]{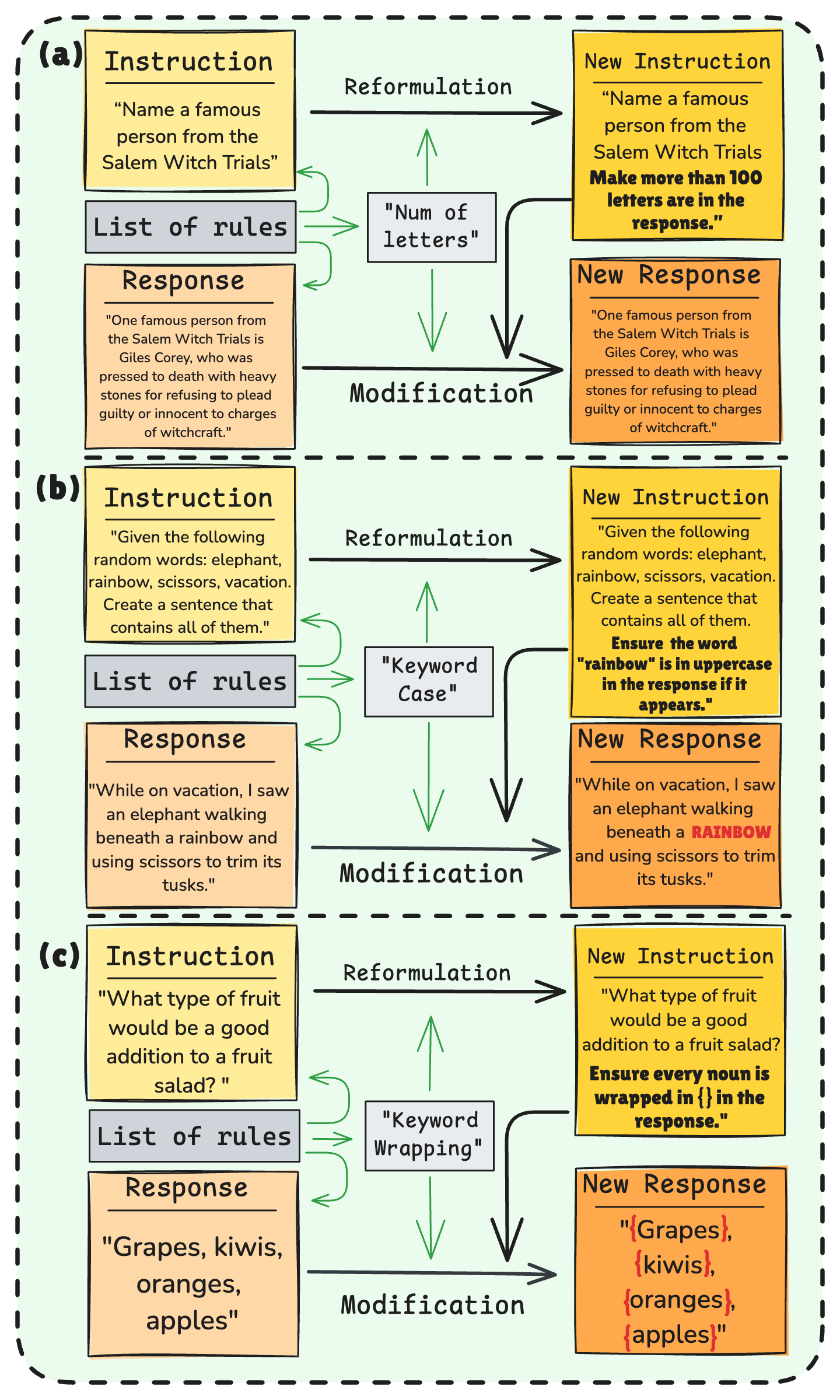}
    \caption{Examples of our data recycling workflows. (a), (b) and (c) select different predefined rules to modify the original data to fulfill constraints on the complexity or format of the response. The differences in new responses are highlighted in \textcolor{red}{red}, the example in (a) has already satisfied the appended constraint, thus the response is kept unchanged. \looseness-1}
    \vspace{-5mm}
    \label{fig:toy_example}
\end{figure}


%% file: table_main_1.tex
\begin{table*}[!ht]
\centering
\scalebox{0.62}{
\begin{tabular}{l|l|c|cccc|c|c|cccc}
\toprule
\multirow{2}{*}{\textbf{Model}}& \multirow{2}{*}{\textbf{Dataset}} & \multirow{2}{*}{\textbf{Method}}  & \multicolumn{4}{c|}{\textbf{Instruction Following Eval}} & \textbf{Pair-wise }& \multicolumn{5}{c}{\textbf{Open LLM Leaderboard} }  \\
 & & & Prompt (S) & Inst (S) & Prompt (L) & Inst (L) & \textbf{Winning Score}  & \textbf{Average} & A & H & M & T \\
\midrule
\multirow{10}{*}{\textbf{Mistral-7B}}& \multirow{2}{*}{\textbf{Alpaca-GPT4}} & Baseline & 32.72 & 42.45 & 35.67 & 45.44 & 1.000 & \textbf{61.24} & 56.23 & 81.07 & 56.22 & 51.42 \\
& & Ours & \textbf{39.56} & \textbf{51.44} & \textbf{43.44} & \textbf{55.40} & \textbf{1.044} & 61.05 & 56.66 & 80.50 & 57.24 & 49.81 \\
\cmidrule{2-13}
& \multirow{2}{*}{\textbf{Alpaca}} & Baseline & 33.64 & 44.60 & 36.23 & 47.96 & 1.000 & 55.15 & 51.96 & 74.61 & 52.85 & 41.20 \\
& & Ours & \textbf{35.12} & \textbf{46.76} & \textbf{37.89} & \textbf{49.52} & \textbf{1.158} & \textbf{56.21} & 54.61 & 77.70 & 54.54 & 38.00 \\
\cmidrule{2-13}
& \multirow{2}{*}{\textbf{Wizard-70k}} & Baseline & 37.34 & 48.68 & 40.85 & 52.04 & 1.000 & 59.38 & 54.61 & 79.96 & 55.68 & 47.27 \\
& & Ours & \textbf{46.77} & \textbf{57.07} & \textbf{49.17} & \textbf{59.71} & \textbf{1.168} & \textbf{59.75} & 55.38 & 80.75 & 55.59 & 47.27 \\
\cmidrule{2-13}
& \multirow{2}{*}{\textbf{Recycled Wizard}} & Baseline & 30.87 & 42.41 & 35.86 & 46.40 & \textbf{1.000} & 59.61 & 54.10 & 77.85 & 57.61 & 48.87 \\
& & Ours & \textbf{39.56} & \textbf{51.08} & \textbf{45.10} & \textbf{56.24} & 0.987 & \textbf{60.43} & 56.66 & 78.01 & 58.61 & 48.43 \\
\cmidrule{2-13}
& \multirow{2}{*}{\textbf{DEITA 6K}} & Baseline & 41.22 & 51.08 & 44.55 & 54.92 & 1.000 & 64.82 & 60.41 & 82.52 & 61.57 & 54.76 \\
& & Ours & \textbf{42.14} & \textbf{52.28} & \textbf{46.77} & \textbf{56.59} & \textbf{1.010} & \textbf{65.43} & 61.86 & 82.71 & 62.66 & 54.49 \\
\midrule
\multirow{6}{*}{\textbf{Llama2-7B}}& \multirow{2}{*}{\textbf{Alpaca-GPT4}} & Baseline & 26.25 & 36.33 & 30.31 & 40.29 & 1.000 & 58.71 & 54.69 & 80.05 & 47.89 & 52.21 \\
& & Ours & \textbf{32.35} & \textbf{42.09} & \textbf{35.30} & \textbf{45.56} & \textbf{1.070} & \textbf{59.77} & 56.74 & 80.67 & 48.45 & 53.21 \\
\cmidrule{2-13}
& \multirow{2}{*}{\textbf{Alpaca}} & Baseline& 31.42 & 40.77 & 33.46 & 43.17 & 1.000 & \textbf{55.25} & 54.35 & 78.65 & 47.02 & 40.98 \\
& & Ours & \textbf{34.38} & \textbf{44.36} & \textbf{37.34} & \textbf{47.36} & \textbf{1.023} & 55.24 & 54.61 & 78.76 & 46.17 & 41.42 \\
\cmidrule{2-13}
& \multirow{2}{*}{\textbf{Wizard-70k}} & Baseline & 31.24 & 44.24 & 35.49 & 48.68 & 1.000 & 57.09 & 54.18 & 79.25 & 46.93 & 48.02 \\
& & Ours & \textbf{38.82} & \textbf{50.12} & \textbf{42.33} & \textbf{53.48} & \textbf{1.087} & \textbf{57.25} & 55.20 & 79.81 & 46.61 & 47.38 \\
\midrule
\multirow{6}{*}{\textbf{Llama2-13B}}& \multirow{2}{*}{\textbf{Alpaca-GPT4}} & Baseline & 32.90 & 44.60 & 36.23 & 48.08 & \textbf{1.000} & 61.47 & 58.70 & 83.12 & 54.13 & 49.92  \\
& & Ours & \textbf{36.60} & \textbf{47.00} & \textbf{37.89} & \textbf{49.28} & 0.977 & \textbf{61.96} & 59.47 & 82.88 & 53.98 & 51.52 \\
\cmidrule{2-13}
& \multirow{2}{*}{\textbf{Alpaca}} & Baseline & 34.94 & 44.36 & 36.41 & 46.29 & \textbf{1.000} & \textbf{57.63} & 57.25 & 81.23 & 54.13 & 37.91  \\
& & Ours & \textbf{36.04} & \textbf{48.20} & \textbf{41.22} & \textbf{52.88} & 0.977 & 57.16 & 57.17 & 81.11 & 52.70 & 37.65 \\
\cmidrule{2-13}
& \multirow{2}{*}{\textbf{Wizard-70k}} & Baseline & 43.07 & 53.84 & 46.40 & 57.67 & 1.000 & \textbf{61.24} & 57.04 & 83.39 & 55.76 & 48.78  \\
& & Ours & \textbf{45.47} & \textbf{58.15} & \textbf{50.09} & \textbf{61.99} & \textbf{1.010} & 60.84 & 58.28 & 82.37 & 54.35 & 48.36 \\

\bottomrule
\end{tabular}
}
\caption{\textbf{Main Results.} Evaluation on the Instruction Following Eval, Pair-wise Comparison Winning Score, and the Open LLM Leaderboard. We compare \ours with Baseline for finetuning three base models on several different instruction tuning datasets. 
\textit{Baseline} -- models trained with the original dataset; \textit{Ours} -- models trained with \ours-recycled datasets; 
\textit{Prompt} -- Prompt-level accuracy; \textit{Inst} -- Instruction-level accuracy; \textit{S} and \textit{L} represent Strict and Loose versions. \textit{A}, \textit{H}, \textit{M}, and \textit{T} denote ARC, HellaSwag, MMLU, and TruthfulQA. 
}
\label{tbl:main_1}
\vspace{-1em}
\end{table*}

%% file: table_ablation.tex
\begin{table}[!t]
\centering
\scalebox{0.75}{
\begin{tabular}{l|ccc}
\toprule
\multicolumn{1}{l|}{Evaluation Metrics} & \textbf{IF Eval} & \textbf{Pair-wise} & \textbf{Open LLM}  \\
\midrule
Baseline & 39.07 & 1.000 & 61.24 \\
\midrule
Single Temp. & 42.62 & 0.987 & \textbf{61.43} \\
Diverse Temp. (*) & \textbf{47.46} & \textbf{1.044} & 61.05 \\
\midrule
Max Rule = 1 & 46.14 & \textbf{1.168} & 61.22 \\
Max Rule = 2 & 47.09 & 1.117 & 61.15 \\
Max Rule = 3 (*)  & \textbf{47.46} & 1.044 & 61.05 \\
Max Rule = 4 & 46.89 & 1.003 & \textbf{61.55} \\
Max Rule = 5 & 44.36 & 1.013 & 60.06 \\
\midrule
Aug Rate = 0.1 & 41.72 & 1.020 & \textbf{61.34} \\
Aug Rate = 0.3 & 42.08 & 1.037 & 61.20 \\
Aug Rate = 0.5 & 46.55 & \textbf{1.111} & 61.31 \\
Aug Rate = 0.7 & 46.23 & \textbf{1.111} & 61.18 \\
Aug Rate = 0.9 (*)  & \textbf{47.46} & 1.044 & 61.05 \\
\bottomrule
\end{tabular}
}
\caption{ \textbf{Ablation Study.} ``(*)'' represents default. \looseness-1
}
\vspace{-8mm}
\label{tbl:ablation}

\end{table}

%% file: appendix_experimental.tex
\section{Experimental Setup}
\label{sec:experimental}

\subsection{Implementation Details}
We utilize the prompt and code base from Vicuna \cite{vicuna2023} and flash attention \cite{dao2022flashattention} for all our experiments.

The Adam optimizer \cite{kingma2017adam} is utilized with the batch size of $128$ and with the max token length of $2048$. 
For training on Llama2-7B and Llama2-13B \cite{touvron2023llama2}, the maximum learning rate is set to $2\times10^{-5}$ with a warmup rate of $0.03$ for $3$ epochs. 
For training on Mistral-7B \cite{jiang2023mistral}, the maximum learning rate is set to $1\times10^{-5}$ with a warmup rate of $0.1$ for $2$ epochs.
When utilizing our method, we run the augmentation process $3$/$2$ times to simulate the epochs of training. These augmented data are then mixed together and used for training $1$ epoch. All other configurations are kept the same as the baselines. 
\looseness-1

\subsection{Training Datasets}


We utilize 5 SFT datasets to evaluate the effectiveness of our method:

\noindent
\textbf{Alpaca dataset} \cite{alpaca}: This dataset consists of 52,000 instruction-following samples created using the self-instruct paradigm \cite{wang-etal-2023-self-instruct} and OpenAI's text-davinci-003 model. Characterized as a classical dataset with moderate-quality attributes, it serves as the fundamental validation. 

\noindent
\textbf{Alpaca-GPT4 dataset} \cite{peng2023instruction}: This dataset is an enhanced Alpaca dataset that includes responses generated by GPT-4.

\noindent
\textbf{WizardLM dataset} \cite{xu2023wizardlm}: This dataset is generated by the novel Evol-Instruct method, which utilizes ChatGPT-3.5 to rewrite instructions step by step into more complex ones and generate the corresponding responses. We utilize the 70k version in our method, which comprises 70,000 high-quality SFT samples.

\noindent
\textbf{Recycled WizardLM Dataset} \cite{Li2023ReflectionTuningDR}: This dataset is an improved version of the WizardLM dataset, by utilizing the Reflection-Tuning method. 
In the Reflection-Tuning, the initial dataset undergoes two main phases: Reflection on Instruction and Reflection on Response.
In the first phase, specific criteria are carefully curated to evaluate and refine the initial instructions. During the second phase, responses are thoroughly examined and improved to align with the refined instructions. This process generates a dataset with superior quality compared to the original dataset.

\noindent
\textbf{DEITA dataset} \cite{liu2023makes}: This dataset leverages the DEITA (Data-Efficient Instruction Tuning for Alignment) method to select high-quality data from a pool comprised of several high-quality datasets, such as WizardLM and Alpaca. DEITA employs a score-first, diversity-aware data selection strategy to optimize the selection process. This strategy uses a GPT-as-a-judge scoring system that combines complexity and quality in a practical and straightforward manner. The scores are incorporated with the diversity-based selection, ensuring that all the data maintains high standards of complexity, quality, and diversity.

\subsection{Evaluation Metrics}

We employ 3 commonly accepted metrics for the evaluation, including \textbf{IFEval }(Instruction-Following Eval), \textbf{Pair-wise Comparison}, and \textbf{Open LLM Leaderboard}.

\textbf{IFEval} \cite{zhou2023instructionfollowing} is the primary evaluation metric employed in our study due to its compatibility with our motivation. 
It focuses on evaluating how LLMs follow various additional constraints, such as specifying a word count or requiring the inclusion of certain keywords a specified number of times. 
To avoid the utilization of LLMs during evaluation, it proposes 25 distinct types of verifiable instructions. There are 541 prompts in total and each of them incorporates one or more of these verifiable instructions, ensuring the comprehensiveness of the evaluation.
IFEval serves as a great inspiration for our method, and there exist semantical overlappings between their verifiable instructions and our rules. However, \textbf{the specific prompts used in IFEval are kept unknown in the construction of our rule templates, avoiding potential template leakage}. Moreover, IFEval only needs to verify the responses, while our method needs to modify the responses for the training, which pushes this process a step further. 
Consequently, this benchmark not only facilitates a comprehensive comparison but also provides valuable insights that align with our purpose.

\textbf{Pair-wise Comparison} involves evaluating responses from LLMs like GPT-4, especially in open-domain contexts. This method has shown a notable alignment with human assessments, providing a credible evaluative foundation~\cite{zheng2024judging, li2023alpacaeval, sottana2023evaluation}. We utilize test instruction sets from WizardLM~\cite{xu2023wizardlm} and Vicuna~\cite{chiang2023vicuna}, comprising 218 and 80 diverse human-curated instructions, respectively. Following the framework by \cite{li2024quantity, Li2024SuperfilteringWD}, we prompt judging LLM to rate responses on a scale from 1 to 10 across multiple dimensions such as relevance and accuracy. To mitigate positional bias~\cite{ko2020look, wang2023large}, responses are presented in two distinct sequences. Evaluation outcomes are categorized into the 'win-tie-lose' system for each instruction. One model's response can be noted as Win/Lose only if it has a Higher/Lower score in both directions. The specific evaluation prompt is presented in the Appendix \ref{sec:eval}.

\textbf{Open LLM Leaderboard} employs Eval Harness \cite{eval-harness} to systematically assess generative language models through diverse tasks. Key benchmarks include ARC~\cite{clark2018think}, HellaSwag~\cite{zellers2019hellaswag}, MMLU~\cite{hendrycks2020measuring}, and TruthfulQA~\cite{lin2021truthfulqa}, collectively evaluating reasoning abilities, common-sense knowledge, and factual accuracy. This approach highlights LLM strengths and areas for improvement in critical aspects of language understanding and generation~\cite{gao2021framework}.

%% file: appendix_examples.tex
\section{Examples}
\label{sec:examples}

Examples of multi-rule augmentation are shown in Figure~\ref{complex_example}.

\input{table_examples}

%% file: table_examples.tex
\begin{figure*}[!h]
  
  \rule{\textwidth}{1.5pt} 
        
            \textcolor{teal}{[Instruction]} \\
             \textcolor{violet}{Ensure the 2nd paragraph is enclosed in "-| |-" in the response.}\\
             {Write a story of no more than 150 words about a knight who defends their kingdom from a dragon.} \\
             \textcolor{violet}{Make sure the 2nd sentence is enclosed within "\$ \$", the 4th sentence is enclosed within "\$\$\$ \$", the 6th sentence is enclosed within "\#\# \#", the 8th sentence is enclosed within "*** *" in the response.}
            
            \textcolor{teal}{[Response]} \\
            In the kingdom of Arvania, there lived Sir Cedric, a valiant knight, sworn to defend the land against any threat. \textcolor{red}{\$}One day, the kingdom came under attack from a fire-breathing dragon, terrorizing the people and burning homes.\textcolor{red}{\$} The king called on Sir Cedric, charging him to defeat the dragon and restore peace to the land.\\
            \textcolor{red}{\$\$\$-|}Wielding his sword and shield, Sir Cedric rode out to face the dragon, fearlessly engaging it in combat.\textcolor{red}{\$} The battle was fierce, with the dragon's flames clashing against the knight's steel. \textcolor{red}{\#\#}But Sir Cedric was determined and unrelenting, his bravery and skill unmatched.\textcolor{red}{|- \#} \\
            Finally, with a mighty blow, Sir Cedric struck the dragon down, ending its reign of terror. \textcolor{red}{***}The people of Arvania rejoiced, celebrating the bravery of their hero.\textcolor{red}{*} And Sir Cedric returned in triumph, hailed as the dragon-slaying knight.

            \vspace{-0.2cm}
        
            \rule{\textwidth}{1pt} 

            \textcolor{teal}{[Instruction]} \\
            \textcolor{violet}{Ensure the 1st sentence is in uppercase, the 2nd sentence is in uppercase, the 3rd sentence is in uppercase in the response.} \\
            \textcolor{violet}{Keep the word "way" bounded by "@ @" in the response if it appears.}\\
            {Describe an example of a time you used to influence in a positive way} \\
            \textcolor{violet}{Keep the 2nd sentence wrapped in "[[ ]]", the 4th sentence wrapped in "*** ***", and the 5th sentence wrapped in "\# \#" in the response.}

            \textcolor{teal}{[Response]} \\
            I RECENTLY HAD A TEAM PROJECT AT WORK WHERE I HAD TO INFLUENCE MY TEAM MEMBERS TO COMPLETE THE TASK IN THE MOST EFFICIENT \textcolor{red}{@}WAY\textcolor{red}{@} POSSIBLE. \textcolor{red}{[[}I IDENTIFIED AREAS WHERE OUR TEAM COULD IMPROVE, AND I CONSTANTLY COMMUNICATED AND ENCOURAGED MY TEAMMATES TO STAY FOCUSED AND ON TRACK.\textcolor{red}{]]} I ALSO SOUGHT FEEDBACK ON HOW WE COULD IMPROVE THE PROCESS. \textcolor{red}{***}Through my efforts, I managed to motivate my team to complete the project ahead of schedule while also maintaining high-quality work.\textcolor{red}{***} \textcolor{red}{\#}This helped to boost morale and foster collaboration between our team members.\textcolor{red}{\#} Everyone was very satisfied with the resulting outcome.

        \rule{\textwidth}{0.8pt} 
\caption{Examples with multiple rules selected and implemented. The randomly generated rule-instructions are colored in \textcolor{violet}{violet}. The upper example is augmented by $2$ different rules (Paragraph Wrapping, and Sentence Wrapping); the bottom example is augmented by $3$ different rules (Sentence Case, Keyword Wrapping, and Sentence Wrapping). The format differences in new responses are highlighted in \textcolor{red}{red}. }
\label{complex_example} 
\end{figure*}

%% file: appendix_sub.tex
\section{Detailed Sub-Category Analysis}
\label{sec:sub}

In this section, detailed comparisons between our models and baseline models are provided in Table \ref{tbl:sub} for further analysis of the effects of our method. 
It is worth noting that \textbf{the specific prompts used in IFEval are kept unknown in the construction of our rule templates, avoiding potential overfitting to the templates.}
\looseness-1

Models trained with our recycled data outperform the baseline model consistently in ``Case'', ``Combination'', ``Punctuation'', and ``Start End'' categories, which are partially well-recycled by our method. 
In the ``Length'' category, our models are only slightly better than the baseline model although our recycling method contains this kind of constraints. After further investigation, we find the performance in this category is mainly influenced by the original data length distributions. Since our method does not introduce more data samples, thus not able to improve the performance dramatically, but the better performance indeed provides the model with a better understanding of response length. 
Interestingly, the performance in the ``Language'' category also shows consistent improvement although we do not introduce any more data. Considering the consistency between this performance and the Pair-wise comparison performance, we hypothesize this improvement is caused by the better general instruction-following abilities provided by our method. 

The performance in the ``Content'' category presents one of the limitations of our Rule-based Recycling method, without utilizing other models or human experts to rewrite the instruction and response, it's hard for our method to modify the content of the existing response. The performances of our models in the ``Json'' and ``Keywords'' categories are merely slightly lower, which is mainly affected by the diversity of original training datasets. 

Comparing the performance changes across augmentation rates, LLMs obtain better performances when the augmentation rate is higher, except for ``Case'', indicating the easiness of case-related constraints for LLMs to understand and learn. 
\looseness-1

\input{table_sub}

%% file: table_sub.tex
\begin{table*}[t]
\centering
\scalebox{0.8}{
\begin{tabular}{l|ccccccccc}
\toprule
\textbf{Sub-Category} & \textbf{Case} & \textbf{Combination} & \textbf{Content} & \textbf{Json} & \textbf{Keywords} & \textbf{Language} & \textbf{Length} & \textbf{Punctuation} & \textbf{Start End} \\
\midrule
\midrule
Baseline (Strict) & 22.47 & 16.92 & \textbf{75.47} & \textbf{63.06} & \textbf{44.79} & 58.06 & 31.47 & 12.12 & 58.21 \\
\midrule
Aug Rate = 0.1 & \textbf{78.65} & 63.08 & 45.28 & 59.24 & 33.13 & 74.19 & 32.17 & 30.30 & 79.10 \\
Aug Rate = 0.3 & 76.40 & 58.46 & 45.28 & 45.86 & 30.06 & 58.06 & \textbf{35.66} & 22.73 & 71.64 \\
Aug Rate = 0.5 & 70.79 & 69.23 & 49.06 & 45.22 & 39.88 & \textbf{77.42} & 28.67 & 13.64 & 61.19 \\
Aug Rate = 0.7 & 74.16 & 61.54 & 43.40 & 47.77 & 33.13 & 51.61 & 30.77 & \textbf{77.27} & 79.10 \\
Aug Rate = 0.9 & 67.42 & \textbf{75.38} & 47.17 & 47.13 & 34.97 & 61.29 & 31.47 & 66.67 & \textbf{82.09} \\
\midrule
\midrule
Baseline (Loose) & 24.72 & 20.00 & \textbf{75.47} & \textbf{66.88} & \textbf{48.47} & 64.52 & 37.06 & 15.15 & 58.21 \\
\midrule
Aug Rate = 0.1 & \textbf{79.78} & 69.23 & 45.28 & 61.15 & 37.42 & 74.19 & 36.36 & 40.91 & 80.60 \\
Aug Rate = 0.3 & 77.53 & 66.15 & 45.28 & 47.13 & 36.81 & 64.52 & 39.16 & 25.76 & 76.12 \\
Aug Rate = 0.5 & 77.53 & 70.77 & 49.06 & 45.86 & 42.94 & \textbf{80.65} & 32.17 & 15.15 & 65.67 \\
Aug Rate = 0.7 & 77.53 & 70.77 & 43.40 & 48.41 & 35.58 & 58.06 & 35.66 & \textbf{84.85} & 80.60 \\
Aug Rate = 0.9 & 70.79 & \textbf{80.00} & 47.17 & 48.41 & 40.49 & 67.74 & \textbf{39.16} & 69.70 & \textbf{85.07} \\

\bottomrule
\end{tabular}
}
\caption{
Sub-category performance on IFEval benchmark of Mistral-7B finetuned with \ours-augmented Alpaca-GPT4 data. The top section represents the performance by the strict criterion while the bottom represents the loose. \looseness-1
}
\label{tbl:sub}
\vspace{-2mm}
\end{table*}

%% file: appendix_evaluation.tex
\section{Evaluation Metrics}
\label{sec:eval}
The prompt for pair-wise comparison is shown in Figure \ref{eva_prompt}.

\begin{figure}[h]
  \centering
  \parbox{0.48\textwidth}{
        \rule{0.48\textwidth}{1.5pt} 
        Prompt for Performance Evaluation \\
        \rule{0.48\textwidth}{0.8pt} 
        \textbf{System Prompt} \\
        You are a helpful and precise assistant for checking the quality of the answer. \\

        \textbf{User Prompt} \\
        \text{[Question]}\\
        \textit{Question}\\
        \text{[The Start of Assistant 2's Answer]}\\
        \textit{Answer 2}\\
        \text{[The End of Assistant 2's Answer]}\\
        \text{[The Start of Assistant 2's Answer]}\\
        \textit{Answer 2}\\
        \text{[The End of Assistant 2's Answer]}\\

        We would like to request your feedback on the performance of two AI assistants in response to the user question displayed above. \\
        Please rate the helpfulness, relevance, accuracy, level of details of their responses. Each assistant receives an overall score on a scale of 1 to 10, where a higher score indicates better overall performance. \\
        Please first output a single line containing only two values indicating the scores for Assistant 1 and 2, respectively. The two scores are separated by a space. In the subsequent line, please provide a comprehensive explanation of your evaluation, avoiding any potential bias and ensuring that the order in which the responses were presented does not affect your judgment.

        \rule{0.48\textwidth}{0.8pt} 

  }
\caption{
The prompt we used to request GPT4 to evaluate the responses. 
} 
\label{eva_prompt} 
\end{figure}

%% file: appendix_rules.tex
\section{Predefined Rules}
\label{sec:rules}

\input{table_rules}

In this section, we will dive into the predefined rules describing each constraint specifically. 

\noindent
\textbf{Keyword Appearance} simulates the scenario where specific keywords are required to appear in the responses. 
In this rule, several non-stop words are randomly selected from the original data sample and used as the desired characteristics. 
The placeholder {\textit{Keyword}} in the constraint template will be replaced by the sampled keyword as the rule-instruction. 
This process can be repeated to simulate the constraints on multiple keywords. 
The augmented instruction is the concatenation of the original instruction and rule-instruction. 
The original response does not need to be modified in this rule and is directly used as the augmented response.

\noindent
\textbf{Keyword Frequency} simulates controlling the frequency of specific keywords in generated responses. 
In this rule, several non-stop words and their frequencies are randomly sampled and used as the desired characteristics. 
There are three random sub-situations in the rule: More, Less, or Equal. 
In the ``Equal'' situation, the placeholders \{\textit{N}\} and \{\textit{Keyword}\} will be directly replaced by the sampled keyword and its frequency. 
In the ``More'' or ``Less'' situations, a small random number \(x\) will be randomly generated to adjust the keyword frequency to meet the desired constraint template, such as ``Ensure there are more than \{\textit{N} - x\} \{\textit{Keyword}\}'' or ``Ensure there are fewer than \{\textit{N} + x\} \{\textit{Keyword}\}.'' 
This process can be repeated to simulate constraints on multiple keywords. 
The augmented instruction is the concatenation of the original instruction and rule-instruction. 
The original response does not need to be modified in this rule and is directly used as the augmented response.

\noindent
\textbf{Num of Adjectives} simulates controlling the total number of adjectives in generated responses. 
In this rule, the adjectives in the original response are identified and counted using part-of-speech tagging (POS). 
There are three possible sub-situations in the rule: More, Less, or Exact. 
In the ``Exact'' situation, the placeholder \{\textit{N}\} will be replaced by the number of adjectives. 
In the ``More'' or ``Less'' situations, a small random number \(x\) is randomly generated to adjust {\textit{N}} to meet the constraints, such as ``Ensure the response has more than \{\textit{N} - x\} adjectives'' or ``Ensure the response has fewer than \{\textit{N} + x\} adjectives.'' 
This process can only be used once for each sample. 
The augmented instruction is the concatenation of the original instruction and rule-instruction. 
The original response does not need to be modified in this rule and is directly used as the augmented response.

\noindent
\textbf{Num of Nouns} simulates controlling the number of nouns in generated responses, similar to the ``Num of Adjectives''.

\noindent
\textbf{Num of Verbs} simulates controlling the number of verbs in generated responses.

\noindent
\textbf{Num of Characters} simulates controlling the number of characters in generated responses.

\noindent
\textbf{Num of Letters} simulates controlling the number of letters in generated responses.

\noindent
\textbf{Num of Words} This rule simulates controlling the number of words in generated responses.

\noindent
\textbf{Num of Sentences} simulates controlling the number of sentences in generated responses. The sentences from the original response are segmented by utilizing dependency parsing.

\noindent
\textbf{Num of Paragraphs} simulates controlling the number of paragraphs in generated responses. The paragraphs from the original response are segmented by regular expressions. 

\noindent
\textbf{Num of Bullets} simulates controlling the number of bullet points in generated responses. The bullet points from the original response are segmented by regular expressions. 

\noindent
\textbf{Instruction Repetition} simulates the scenario where the LLM is requested to repeat the instructions before providing the response. 
This process can be applied only once for each instruction. 
The augmented instruction is the concatenation of the original instruction and rule-instruction. 
The augmented response is the concatenation of repeated original instruction and the response.

\noindent
\textbf{Response Repetition} simulates the scenario where the LLM is requested to repeat the responses several times. 
In this rule, the response is repeated \{\textit{N}\} times, where \{\textit{N}\} is a random number. 
This process can only be applied once per data sample. 
The augmented instruction is the concatenation of the original instruction and rule-instruction. 
The augmented response is the concatenation of \textit{N} identical responses.

\noindent
\textbf{UP Case} simulates requesting the entire response is required in uppercase. 
In this rule, the original response is converted to uppercase format entirely. 
This process can only be used once for each response. 
The augmented instruction is the concatenation of the original instruction and rule-instruction. 
The augmented response is the all-uppercase version of the original response. 

\noindent
\textbf{Low Case} simulates requesting the entire response is required in lowercase. 

\noindent
\textbf{Letter Case} simulates the scenario where specific types of letters in the response are required to be in uppercase. 
In this rule, the specific letter {\textit{x}} is sampled from the response, and all occurrences of this letter in the response are capitalized.
This process can be repeated on different random letters.
The augmented instruction is the concatenation of the original instruction and rule-instruction. 
The augmented response is the original response with all specific letters in uppercase.

\noindent
\textbf{Keyword Case} simulates the scenario where specific keywords in the response are required to be in uppercase. The specific keyword {\textit{Keyword}} is sampled from the response.

\noindent
\textbf{Sentence Case} simulates the scenario where the specific sentences in the response are required to be in uppercase. The index of the sentence {\textit{i}} is randomly selected within the total number of sentences in the response.

\noindent
\textbf{Paragraph Case} simulates the scenario where the specific paragraphs in the response are required to be in uppercase. The index of the paragraph {\textit{i}} is randomly selected within the total number of paragraphs in the response

\noindent
\textbf{All Removal} simulates controlling LLM to ignore the use of punctuation. 
In this rule, the punctuation marks in the response will be removed completely, and the new response will serve as the augmented response. 
This process can only be used once for each sample.

\noindent
\textbf{All Replacement} simulates controlling LLM to replace all the original punctuation with a predefined symbol \{\textit{Symbol}\}. 
In this rule, the punctuation marks in the response will be replaced, and the new response will serve as the augmented response. 
This process can only be used once for each sample.

\noindent
\textbf{Target Removal} simulates the scenario where a specific type of punctuation mark \{\textit{Punctuation}\} in the response is ignored. 
In this rule, a random type of punctuation is identified, and all occurrences of this mark are removed, the new response will serve as the augmented response. 
This process can only be used once for each sample to avoid confusion. 

\noindent
\textbf{Target Replacement} simulates the scenario where a specific type of punctuation mark \{\textit{Punctuation}\} in the response is replaced by a specified symbol \{\textit{Symbol}\}. 
In this rule, a random type of punctuation mark is identified, and all occurrences of this mark will be replaced by a predefined symbol in the response. 
This process can only be used once. 

\noindent
\textbf{Keyword Wrapping} simulates the scenario where specific keywords in the response are required to be wrapped in a specified format. 
In this rule, a randomly chosen \{\textit{Keyword}\} is identified, and all occurrences of this keyword are wrapped in the randomly specified \{\textit{Format}\} in the response. 
This process can be repeated several times on different words with various formats. 
The augmented instruction is the concatenation of the original instruction and rule-instruction. 
The augmented response is the original response with all keywords wrapped in the format.

\noindent
\textbf{Sentence Wrapping} simulates the scenario where specific sentences in the response are required to be wrapped in a specified format. The index of the sentence {\textit{i}} is randomly selected within the total number of sentences in the response.

\noindent
\textbf{Bullet Wrapping} simulates the scenario where specific bullet points in the response are required to be wrapped in a specified format. The index of the bullet point {\textit{i}} is randomly selected within the total number of bullet points in the response.

\noindent
\textbf{Paragraph Wrapping} simulates the scenario where a specific paragraph in the response are required to be wrapped in a specified format. The index of the paragraph {\textit{i}} is randomly selected within the total number of paragraphs in the response.

\noindent
\textbf{Instruction Wrapping} simulates a scenario where the original instruction is required to be repeated in a specified format before providing the response. 
In this rule, the original instruction is restated with wrapping in the randomly chosen \{\textit{Format}\} before giving the actual response.
This process can be applied only once for each instruction. 
The augmented instruction is the concatenation of the original instruction and rule-instruction. 
The augmented response is the concatenation of the original instruction wrapped in the specific format and the response.

\noindent
\textbf{Response Wrapping} simulates a scenario where the response wrapped in the specific format is required to be repeated several times. 
In this rule, the response is repeated \{\textit{N}\} times wrapeed in the specified \{\textit{Format}\}, with \{\textit{N}\} and \{\textit{Format}\} being randomly selected. 
This process can be applied only once. 
The augmented instruction is the concatenation of the original instruction and rule-instruction. 
The augmented response is the concatenation of the repeated response wrapped in the format.

%% file: table_rules.tex
\begin{table*}[!ht]
\centering
\small 
\scalebox{1}{ 
\begin{tabular}{p{0.35\columnwidth}|p{0.35\columnwidth}|p{1.2\columnwidth}}
\toprule
\textbf{Rule Type} & \textbf{Rule Name} & \textbf{Example of an Instruction Template for the Rule} \\
\midrule
Keyword Frequency & Keyword Appearance & Ensure \{\textit{Keyword}\} is in the response. \\
\midrule
Keyword Frequency & Keyword Frequency & Ensure there are more/less/exact \{\textit{N}\} \{\textit{Keyword}\} in the response. \\\midrule
Number Constraint & Num of Adjectives & Ensure the response has more/less/exact \{\textit{N}\} adjectives. \\\midrule
Number Constraint & Num of Nouns & Ensure the response has more/less/exact \{\textit{N}\} nouns. \\\midrule
Number Constraint & Num of Verbs & Ensure the response has more/less/exact \{\textit{N}\} verbs. \\\midrule
Number Constraint & Num of Characters & Ensure the response has more/less/exact \{\textit{N}\} characters. \\\midrule
Number Constraint & Num of Letters & Ensure the response has more/less/exact \{\textit{N}\} letters. \\\midrule
Number Constraint & Num of Words & Ensure the response has more/less/exact \{\textit{N}\} words. \\\midrule
Number Constraint & Num of Sentences & Ensure the response has more/less/exact \{\textit{N}\} sentences. \\\midrule
Number Constraint & Num of Paragraphs & Ensure the response has more/less/exact \{\textit{N}\} paragraphs. \\\midrule
Number Constraint & Num of Bullets & Ensure the response has more/less/exact \{\textit{N}\} bullet points. \\\midrule
Repetition & Instruction Repetition & Repeat the instruction before providing the response. \\\midrule
Repetition & Response Repetition & Repeat the response \{\textit{N}\} times. \\\midrule
Case All &  Up Case & Ensure the response is all in upper case. \\\midrule
Case All &  Low Case & Ensure the response is all in lowercase. \\\midrule
Case Target &  Letter Case & Ensure all the letters \{\textit{x}\} in the response are in uppercase. \\\midrule
Case Target &  Keyword Case & Ensure all the word \{\textit{Keyword}\} in the response are in uppercase. \\\midrule
Case Target &  Sentence Case & Ensure \{\textit{i}\}-th sentence in the response is in uppercase. \\\midrule
Case Target &  Paragraph Case& Ensure \{\textit{i}\}-th paragraph in the response is in uppercase. \\\midrule
Punctuation All & All Removal  &  Ignore all punctuation in the response. \\\midrule
Punctuation All & All Replacement & Use \{\textit{Symbol}\} to replace all punctuation in the response. \\\midrule
Punctuation Target &  Target Removal &  Ignore \{\textit{Punctuation}\} punctuation in the response. \\\midrule
Punctuation Target & Target Replacement & Use \{\textit{Symbol}\} to replace \{\textit{Punctuation}\} in the response. \\\midrule
Format Wrapping &  Keyword Wrapping & Ensure every \{\textit{Keyword}\} is wrapped in \{\textit{Format}\} in the response.  \\\midrule
Format Wrapping &  Sentence Wrapping & Ensure \{\textit{i}\}-th sentence is wrapped in \{\textit{Format}\} in the response.  \\\midrule
Format Wrapping &  Bullet Wrapping & Ensure \{\textit{i}\}-th bullet point is wrapped in \{\textit{Format}\} in the response.  \\\midrule
Format Wrapping &  Paragraph Wrapping & Ensure \{\textit{i}\}-th paragraph is wrapped in \{\textit{Format}\} in the response.  \\\midrule
Formatted Repeating &  Instruction Wrapping & Repeat the instruction in \{\textit{Format}\} before providing the response.  \\\midrule
Formatted Repeating &  Response Wrapping & Repeat the response \{\textit{N}\} times in \{\textit{Format}\}.  \\
\bottomrule
\end{tabular}}
\caption{The list of predefined constraint rules. 
Each rule contains (1) a set of constraint templates that serve as additional rule-instructions, on constraints the LLM should follow, and (2) specified methods that alternately edit the instruction and response to reach an alignment between them. }
\label{tab:list}
\end{table*}